# AN ONLINE STOCHASTIC KERNEL MACHINE FOR ROBUST SIGNAL CLASSIFICATION

*Raghu G. Raj*

U.S. Naval Research Laboratory, Radar Division
Washington D.Cs. 20375, U.S.A.

**ABSTRACT**

We present a novel variation of online kernel machines in which we exploit a consensus based optimization mechanism to guide the evolution of decision functions drawn from a reproducing kernel Hilbert space (RKHS) such that the entire stationary process observed can be efficiently modeled. We derive an efficient classification algorithm based on these principles such that our algorithm reduces to traditional online kernel machines for the special case in which the consensus based optimization mechanism is switched off. We illustrate the inherent label and input noise resistance of our algorithm for the case of online classification; and derive relevant mistake bounds. The resulting algorithm can find numerous applications such as, for example, in target classification by remote sensing platforms wherein the target being classified tends to typically be persistent over the observation interval.

## 1. INTRODUCTION

The design of inference engines—for a variety of applications ranging from enabling robot intelligence to remote sensing systems—is a central aspect of machine learning. Inference engines are typically designed either for batch processing or online learning modes. Whereas in the former setting, learning is based on a random batch of training samples from which a single hypothesis is formulated for prediction, in the online setting the learning algorithm observes a sequence of samples from which predictions are made one at a time.

Online learning is particularly applicable to cases—such as for example distributed learning via wireless networks—where there are no a prior training sets available, where the learning must proceed dynamically from samples acquired on-the-fly, or where the on-board processor memory is limited.

Perhaps the earliest online classification technique is the Perceptron algorithm [1-2] in which a linear decision boundary is updated by a simple additive update rule that pushes the decision boundary in the direction that rectifies the current misclassification. When applied to high dimensional feature spaces, however, the perceptron is susceptible to overfitting owing to the lack of regularization. The incorporation of kernel methods into the online algorithms has been shown to be an effective way to handle this problem, and several variants of which have been proposed in the literature [3-6]. On the other hand online learning can also be framed in terms of a Bayesian update of the posterior distribution [7]. Though such techniques are powerful, they typically involve computationally intensive update rules.

Early work on the theoretical limits of online learning were investigated from a statistical physics point of view [8] (for more modern physics based approaches to machine learning, see [19-20]). Though these techniques benchmark the performance of online algorithms in idealized scenarios they are of limited practical value because they are based on specific parametric statistical models. Furthermore statistical physics based approaches typically require a priori knowledge of parameters such as generalization error which are not knowable in practice. On the other hand the methods of statistical learning theory provide greater insight into the behavior of online algorithms [9], particularly kernel based algorithms in which we are particularly interested.

This paper explores a novel variation of online kernel machines in which we exploit a consensus based *optimization* mechanism to guide the evolution of decision functions drawn from a reproducing kernel Hilbert space (RKHS) that efficiently models the entire observed stationary process. The motivation for this stems from the fact that in practice the target being classified, for example in aerial surveillance applications, typically tends to be persistent over the observation interval. This information can therefore be used as a prior in lending robustness to noise in the labels that are fed back to the online learning system. We demonstrate that our resulting algorithm has superior robustness to input and label noise compared to the vanilla kernel based online classification algorithms.

The rest of this paper is organized as follows. In section 2 we review the basics of online kernel machines, and define the notation that will be used throughout the paper. In section 3 we formulate our novel online stochastic kernel machine (osKM) algorithm from first principles and perform basic mistake bound analysis. In section 4 we apply the osKM algorithm to the problem of online signal classification with input and label noise and compare the performance to a traditional online learning counterpart. Finally we conclude the paper in Section 5 with directions for future work.

This work is supported by ONR NRL via the 6.1 Base Program

## 2. ONLINE KERNEL MACHINES

Let $S_N = \{(x_1, y_1), (x_2, y_2), \ldots, (x_N, y_N)\}$ be a sequence of samples observed by the online learning algorithm, where $x_k$ is the $k^{th}$ sample/feature vector and $y_k$ is the corresponding class label. The goal of a learning system is to infer a hypothesis function $f$ drawn from a space $H$ such that the application of $f$ to unseen samples in the future, drawn from the same distribution as the samples observed thus far, will correctly predict the class label in most cases (i.e. low generalization error). In order for this formalism be to be computationally realizable, more structure needs to be placed onto $H$.

Specifically, in the context of kernel machines the set $H$ is modeled as a RKHS whereby there exists a kernel function $k: \mathcal{X} \times \mathcal{X} \to \mathbb{R}$ such that [9]:
i) $k$ has the reproducing property:
$$\langle f, k(x, \cdot) \rangle = f(x), \forall x \in \mathcal{X}$$
ii) $H$ is the closure of the span of $k(x, \cdot)$ with $x \in \mathcal{X}$

Given this set up, the primary objective is typically to minimize the empirical risk defined as [9]:

$$R_{emp}[f; S_N] = \frac{1}{N} \sum_{n=1}^{N} l(f(x_t), y_t) \quad (1)$$

where, $l$ is the loss function between the actual class labels $\{y_t\}$ and the predicted labels $\{f(x_t)\}$. For the case of signal classification the ideal loss function to employ is the so-called 0-1 loss function:

$$l(f(x), y) = \phi_0(yf(x)) \equiv \mathbb{I}[y \neq f(x)] \quad (2)$$

where, $\mathbb{I}$ is the discrete Dirac delta operator. However since $\phi_0$ is non-convex, convex surrogates are employed in practice such as hinge, exponential and logistic loss functions. In this paper we focus on the hinge loss induced soft margin loss function which correspond to support vector (SV) machines [9]:

$$l(f(x), y) = \phi_1(yf(x)) \equiv max\{1 - yf(x), 0\} \quad (3)$$

In order to ensure good generalization ability it is essential to incorporate regularization into (1) which will prevent overfitting of the data. In the context of kernel machines, this is accomplished by minimizing the following modified cost function:

$$R_{reg0}[f; S_N] = \frac{1}{N} \sum_{n=1}^{N} l(f(x_n), y_n) + \Omega(f) \quad (4)$$

where, $\Omega(f)$ is a monotonically increasing function. In this paper we exclusively focus on the case where $\Omega(f) = \|f\|_H^2$ due to analytical tractability, although other norms such as the $l_p$ norm can also be incorporated. It can be shown that regularization controls the complexity of the learned classifier which is proportional to the number of support vectors used by the kernel machine in determining the decision boundary in higher dimensional kernel space [9].

Given the current estimate of the decision function $f_N$ based on the observations $S_N$, the general strategy, in the online setting, is to determine the optimum increment by a (stochastic) steepest gradient update of the cost function (4). Variants of this general strategy have appeared in the literature that differ with regards to the specific structure of the cost function [10], the manner in which the space of previous samples are modeled [6], the specific analytical and numerical constraints placed on the update rules [9], and the manner in which the growth of the support vectors is controlled [5].

## 3. ONLINE STOCHASTIC KERNEL MACHINES
### 3.1. Background

An important consideration in many application is the presence of noise—both input and label noise. For example in sensor network applications consisting of a set of low cost sensors, the sensors typically have relatively high A/D (analog-to-digital) noise figures which manifest in the form of input noise being added to the received signal. Compounding this, label noise naturally arises in applications such as distributed sensing where the decision made by the fusion center (or ad hoc network) is likely to have errors—which when fed back to the sensor nodes can potentially degrade the overall system performance. Thus it is highly desirable to investigate input and label noise resistant methods of online learning [10-12].

In this paper we introduce a novel approach to label and input noise resistant kernel machines wherein we incorporate the notion of persistence of (various notions of) class information over time in a flexible and mathematically elegant manner.

### 3.2. Incorporating Stochastic Structure

Kernel machines are based on the following regularized cost function for signal classification:

$$R_{reg}[f; S_N] = \frac{1}{\tau} \sum_{n=1}^{\tau} l(f(x_n), y_n) + \frac{\lambda}{2} \|f\|_H^2 \quad (5)$$

where, $\|f\|_H = \langle f, f \rangle^{1/2}$, $l$ is a loss function such as the hinge loss, and $(x_n, y_n)$ is a sample where $x_n$ is the $n^{th}$ sample/feature vector and $y_n$ is the corresponding class label

Given data sequence $S_N$, let $\tau_p$ represent the length of the window of consecutive data samples (which we call the *persistence window*), over which the labels are modeled to be stationary. Then the $i^{th}$ realization of the stochastic process $\mathcal{X} = [\mathcal{X}_1, \mathcal{X}_2, \cdots, \mathcal{X}_{\tau_p}]$ is given by:

$$X_i = [x_{\mathcal{I}(i)}, x_{\mathcal{I}(i)-1}, \cdots, x_{\mathcal{I}(i)-\tau_p+1}]^T \quad (6)$$

where: $\mathcal{I}(i)$ is the index of the latest sample within the data sequence $S_N$ corresponding to the $i^{th}$ realization $X_i$. Let $\tilde{X}_i = [X_i^T; X_{i-1}^T; \cdots, X_{i-\tau+1}^T]$ represent the sequence of stochastic realizations of $\mathcal{X}$ within the observation interval $\tau > \tau_d$. Note that, by construction, $\tilde{X}_i(:, k)$ is the data sequence that pertains to the random variable $\mathcal{X}_k$. We further remark that the definitions of the index function $\mathcal{I}(i)$ and persistence window $\tau_p$ are completely arbitrary; its specific instantiation is based on knowledge of the evolution of the stochastic process $\mathcal{X}$.

At each realization of $\mathcal{X}$, our goal is to calculate the corresponding optimum weights $\alpha$ by solving (5). Our general strategy is to formulate (5) in terms of an alternative direction method of multipliers (ADMM) [15] framework as follows:

$$\underset{\alpha,v}{\text{minimize}} \sum_{j=1}^{\tau_p} g_j(\alpha_j^n; \tilde{X}_{\hat{\jmath}(j,n)}) \quad (7)$$
$$\text{subject to}$$
$$K_j^{n,\tau} \alpha_j^n - v = 0$$

Where:
$$g_j(\alpha_j^n; \tilde{X}_i) =$$
$$\frac{1}{\tau}\sum_{l=1}^{\tau} \phi\left(\sum_{m=1}^{\tau} \alpha_j^n(m) y(m) y(l) k\left(\tilde{X}_{\hat{\jmath}(j,n)}(l,j), \tilde{X}_{\hat{\jmath}(j,n)}(m,j)\right)\right)$$
$$+\frac{\lambda}{2}\sum_{l=1}^{\tau}\sum_{m=1}^{\tau} \alpha_j^n(l)\alpha_j^n(m) y(l) y(m) k\left(\tilde{X}_{\hat{\jmath}(j,n)}(l,j), \tilde{X}_{\hat{\jmath}(j,n)}(m,j)\right) \quad (8)$$

$$K_j^{n,\tau} = \widehat{K_j^{n,\tau}} / \|\widehat{K_j^{n,\tau}}\|_F \quad (9)$$

$$\widehat{K_j^{n,\tau}} = [k_1, k_2, \cdots, k_{\tau_e}]^T \quad (10)$$

$$k_i = [k(x_{n-i+1}, x_{n-j+\tau+k+1})]_{k=1}^{\tau} \quad (11)$$

And such that:

$\alpha_j^n \in \mathbb{R}^\tau$, $v \in \mathbb{R}^\tau$, $K_j^{n,\tau} \in \mathbb{R}^{\tau_e \times \tau}$

$\hat{\jmath}(j,n) = \mathcal{I}(n - j + 1)$

$k(.,.)$ is the kernel matrix operator

$\phi$ is the convex surrogate of the 0-1 loss function

We remark that that $g_j$ in (8) is the standard kernel based cost function in (4). In the above notation, the index $n$ denotes the current time sample, $\tau$ denotes the time of the observation window (with respect to time $n$), $\tau_p$ is the size of the persistence window, and $\tau_e$ is the window over which the kernel is evaluated (such that $\tau_p \leq \tau_e \leq \tau$).

Intuitively, the term $v$ captures the (soft-version of the) class predictions of the data samples. Thus minimizing the collaborative optimization criterion tends to enforce uniformity among the class predictions within the persistence window. Thus another novel aspect of our formulation is that the consensus optimization power of the ADMM framework is leveraged not for distributed processing applications (as in [16]), but rather to enforce the weights α to efficiently model the entire stationary process.

The optimization of (7) follows the ADMM methodology whereby the variables to be optimized are split into 2 parts $\alpha$ and $v$—which are solved, respectively, by 2 sub-problems that both involve optimization of a quadratic augmented Lagrangian [13]:

$$\alpha_j^{n+1} = \underset{\alpha}{\text{argmin}} \begin{pmatrix} g_j(\alpha_j^n; \tilde{X}_{\hat{\jmath}(j,n)}) \\ +(\frac{\rho}{2})\|v^n - K_j^{n,\tau}\alpha_j^n + u_j^n\|_2^2 \end{pmatrix} \quad (12a)$$

$$v^{n+1} = \underset{\alpha}{\text{argmin}} \left(\frac{1}{\tau_p}\sum_{j=1}^{\tau_p}\|v - K_j^{n,\tau}\alpha_j^{n+1} + u_j^n\|_2^2\right) \quad (12b)$$

$$u_j^{n+1} = u_j^n + K_j^{n,\tau}\alpha_j^{n+1} - v^{n+1} \quad (12c)$$

where: $u$ is the scaled dual variable associated with the dual ADMM optimization problem.

It is straightforward to show that the optimization problem (12b) can be solved via the application of the following formula:

$$v^{n+1} = \frac{1}{\tau_p}\sum_{j=1}^{\tau_p}(u_j^n + K\alpha_j^{n+1}) \quad (13)$$

Though (12a) can be optimized via batch-mode like kernel machine optimization equations [9, 13], such an approach can be computationally expensive owing to high dimensionality of the underlying conditional probability densities that are involved. Rather this problem is much better suited to the online setting where it is convenient and efficient to solve it via a stochastic update approach that we described in Section 2. In our case we can show that the update equation (12a) reduces to:

$$\alpha_j^{n+1} = \alpha_j^n - \eta\lambda K_j^{n+1,\tau}\alpha_j^n - \eta \begin{bmatrix} \sum_{l=1}^{n+1} y_l K_{j,l}^T \frac{\partial \phi(\beta_j^l)}{\partial \beta_j^l}\bigg|_{\beta_j^l=c} \\ \sum_{l=1}^{n+1} y_l k(x_{n+1}, x_l) \frac{\partial \phi(\beta_j^l)}{\partial \beta_j^l}\bigg|_{\beta_j^l=c} \end{bmatrix}$$
$$- \rho\left(K_j^T K_j \alpha_j^n + K_j(u_j^n - v^n)\right) \quad (15)$$

Where: $K_{j,l} = K_j^{n+1,\tau}(l, 2:n+1)$,

$c = y_l K_{j,l}\tilde{\alpha}_j^n + y_l \tilde{\alpha}_j^n(n+1)k(x_{n+1}, x_l)$, $\tilde{\alpha}_j^n = \alpha_j^n(2:n+1)$

The next subsection describes the resulting algorithm in greater detail.

### 3.3. The osKM Algorithm

Based on the structural elucidations of the previous section, we can now formally state the osKM algorithm:

*Input data:*
Sequence $S_N = \{(x_1, y_1), (x_2, y_2), \ldots, (x_N, y_N)\}$ where $x_k$ is the kth feature vector and $\tilde{y}_k$ the corresponding noisy label

*Algorithm:*
1) Initialize $f^1 = k(z,.)$, where $z$ is an all-zeros vector (where $f^n$, corresponding to weights $\alpha^n$, represents the optimum hypothesis function at the end of the $n^{th}$ realization of the stochastic process $\mathcal{X}$)
2) Loop (over the index $n$ of stochastic realizations of $\mathcal{X}$)
3) Perform $\tau_p$ on-line updates of the respective

weights $\alpha_j^n$ via the application of stochastic on-line updates via (15)
4) Perform update of the variable $v^n$ via (13)
5) Update the scaled dual variable via (12c)
6) End loop

The added time complexity at each step is due to $\tau_p$ terms in the summation associated with the consensus optimization component together with a constant number ADMM iterations.

## 3.4. Mistake Bound Analysis

*Theorem:* Let $I$ denote the number rounds of the osKM algorithm makes an update when processing a sequence of samples $x_1, x_2, \ldots, x_N \in \mathbb{R}^\tau$. For any $\rho > 0$ and $z \in \mathbb{R}^\tau$ with $\|z\| \leq 1$, consider the vector of $\rho$ − hinge losses incurred by $u$: $L_\rho^{osKM}(z) = \left[\left(1 - \frac{y_t(\langle Q(z), x_n\rangle)}{\rho}\right)_+\right]_{n \in I}$; where: $Q$ is the effective linear projection operator due to the action of (12). Then, the number of updates $M_N = |I|$ made by the osKM algorithm can be bounded as follows:

$$M_N \leq \inf_{\rho>0, \|u\|\leq 1} \left(\frac{\|L_\rho^{osKM}(z)\|_2}{2} + \sqrt{\frac{\|L_\rho^{osKM}(z)\|_2^2}{4} + \frac{tr\{K\}}{\rho}}\right)^2$$

where: $K \equiv K_j^{N,\tau}$

*Proof Sketch.* The proof follows a similar methodology as in [18] except that the projection operator $Q$ serves to sharpen the bound associated with the hinge-loss function □

Therefore, osKM achieves tighter a mistake bound w.r.t. a standard kernel machine due to exploiting additional structure of the stochastic process evolution.

## 4. SIMULATION RESULTS

We consider a binary classification problem for different levels of input and label noise. We draw samples from a collection of curated Synthetic Aperture Radar (SAR) images called the *MSTAR (Moving and Stationary Target Recognition) database* [16]. For our experiments we choose the following two targets classes: BMP2 (Infantry Fighting Vehicle) and BTR-70 (Armored Personnel Carrier). Each image sample is a 64x64 grayscale image. For each of the sample images, we extract Wavelet features [17] (using the 'Reverse biorthogonal wavelet transform')—LH, HL and HH bands—resulting in a 128 dimensional feature vector.

In order to simulate the persistence of target over an observation interval, we vary the ground-truth labels as shown in Figure 1(a) (where labels 1 and -1 represent BMP2 and BTR-70 respectively). This corresponds to the non-stationary case where the target persistence has a short duration with respect to the observation interval. Furthermore, in our simulations the labels are randomized within a periodic non stationary structure above. In particular, the simulations are iterated 100 times in order to obtain confidence intervals.

For all cases that follow the linear kernel was employed and all the parameters were determined by cross-validation. In particular for the simulations below the following parameter values were used: $\lambda = 0.1$, $\rho = 0.1$, $\eta = 0.7$, $\tau = 100$, $\tau_p = 10$, $\tau_e = \tau_p$. In these experiments, we found that as little as three ADMM iterations (per sample image) was sufficient to derive significant performance gain over standard on-linear classification methods.

In Figure 1(b). the performance of the osKM algorithm and a standard on-line kernel based algorithm, called Norma [3], is demonstrated for varying levels of input noise. We remark that this experimental set-up is challenging for on-line methods such as Norma due to the non-stationary structure of the input data stream which accounts for the low performance of the Norma algorithm across the input noise levels. The osKM algorithm, on the other hand, demonstrates robust classification performance across input noise levels.

In Figure 1(c) we demonstrate the performance of osKM and Norma for varying levels of label noise. Here we find that the osKM algorithm outperforms Norma for label noise level less than roughly 30%. If on the other hand, roughly 50% of the labels are flipped then performing osKM renders no advantage over Norma (since in this case the labels are wholly non-informative).

In Figure 2(a-c) we consider the corresponding performance of osKM and Norma for the stationary case where target persistence has a relatively large duration with respect to the observation interval. In this less challenging scenario, we find that the performance of traditional online methods is much closer to osKM. However even in this scenario, osKM has a statistically significant performance advantage over Norma. These results therefore demonstrate the importance of handling the non-stationary structure of labels (in addition to noise considerations) in the performance of an online inference algorithm.

These results demonstrate the effectiveness and power of our novel algorithmic framework for robust on-line signal classification applications.

## 5. SUMMARY

In this paper we have introduced novel processing structures for incorporating a consensus based optimization mechanism into the evolution of decision functions drawn from a RKHS, derived a simple efficient algorithm based on this principle, performed basic mistake bound analysis, and demonstrated superior discrimination performance in the presence of input and label noise compared to its traditional online inference counterpart. Furthermore, our results demonstrate the importance of handling the non-stationary structure of labels (in addition to noise considerations) in the performance of an online inference algorithm. An important direction for future work emanating from this paper is to generalize the algorithm and analysis to data over arbitrary manifolds.

## 6. REFERENCES

[1] F. Rosenblatt. "The perceptron: A probabilistic model for information storage and organization in the brain," *Psychological Review*, 65:386–407, 1958.

[2] M.L. Minsky and S.A. Papert, *Perceptrons*, Cambridge, MA: MIT Press, 1969.

[3] J. Kivinen, A.J. Smola, and R.C. Williamson, "Online Learning with Kernels," *IEEE Transactions on Signal Processing*, Vol. 52, No. 8, August 2004.

[4] C. Gentile, "A new approximate maximal margin classification algorithm," *J. Machine Learning Res.*, vol. 2, pp. 213–242, Dec. 2001.

[5] O. Dekel, S. Shalev-Shwartz, and Y. Singer, "The Forgetron: A Kernel-Based Perceptron on a Budget," *SIAM Journal on Computing*, Vol. 37 Issue 5, pp.1342-1372, January 2008.

[6] F. Orabona, J. Keshet, and B. Caputo, "The projectron: a bounded kernel-based Perceptron," *ICML '08 Proceedings of the 25th international conference on Machine learning*, pp. 720-727, 2008.

[7] M. Opper, "A Bayesian approach to on-line learning," (Published in: *On-line learning in neural networks*, Cambridge University Press), pp. 363 - 378, 1998.

[8] W. Krauth and M. Mezard, "Learning algorithms with optimal stability in neural networks," *J. Phys. A: Math. Gen.*, 20 L745, 1987.

[9] B. Scholkopf and A.J. Smola, *Learning with Kernels: Support Vector Machines, Regularization, Optimization, and Beyond*, MIT Press Cambridge, MA, USA, 2001.

[10] N. Cesa-Bianchi, S. Shalev-Shwartz and Ohad Shamir, "Online Learning of Noisy Data," *IEEE Transactions on Information Theory*, Vol. 57, No. 12, pp.7907-7931, August 2011.

[11] T. Bylander, "Learning linear threshold functions in the presence of classification noise," *In Proc. of the 7th COLT, ACM*, pages 340-347, USA, 1994.

[12] N. Natarajan, I.S. Dhillon, P. Ravikumar, and A. Tewari, "Learning with noisy labels," *Proceedings of the 26th International Conference on Neural Information Processing Systems (NIPS)*, Vol.1, pp. 1196-1204, 2013.

[13] S. Boyd, N. Parikh, E. Chu, B. Peleato, and J. Eckstein, "Distributed optimization and statistical learning via the alternating direction method of multipliers," *Foundation Trends Machine Learning*, 3(1):1–122, 2011.

[14] P. A. Forero, A. Cano, and G. B. Giannakis, "Consensus-based distributed support vector machines," *Journal of Machine Learning Research*, vol. 11, pp. 1663–1707, 2010.

[15] E. Esser, "Applications of Lagrangian-based alternating direction methods and connections to split Bregman," *CAM report*, vol. 9, p. 31, 2009.

[16] "The Airforce Moving and Stationary Target Recognition Database." [Online]. Available: *https://www.sdms.afrl.af.mil/datasets/mstar/*

[17] S.G. Mallat, *A Wavelet Tour of Signal Processing*, 2nd Edition (New York: Academic), 1999.

[18] M. Mohri and A. Rostamizadeh, "Perceptron mistake bounds," *arXiv 1305.0208*, March 2013.

[19] P. Chao, T. Mazaheri, B. Sun, N.B. Weingartner, and Z. Nussinov, "The Stochastic Replica Approach to Machine Learning: Stability and Parameter Optimization," *arXiv:1708.05715 [stat.ML]*, November 2018.

[20] T. Mazaheri, Bo Sun, J. Scher-Zagier, A. S. Thind, D. Magee, P. Ronhovde, T. Lookman, R. Mishra, and Z. Nussinov, "Stochastic replica voting machine prediction of stable cubic and double perovskite materials and binary alloys," *Phys. Rev. Materials* 3, 063802, June 2019.

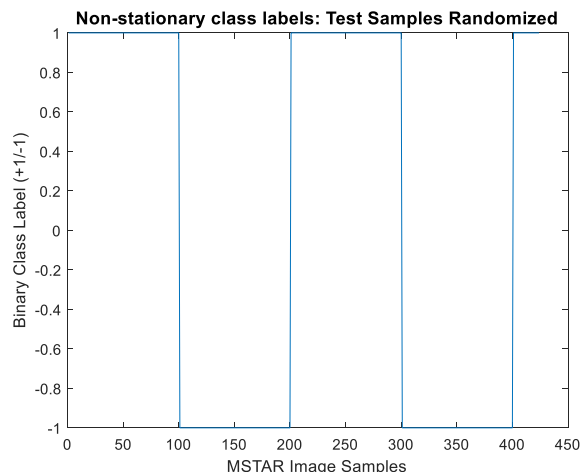

*Figure 1.* Non-stationary ground-truth labels

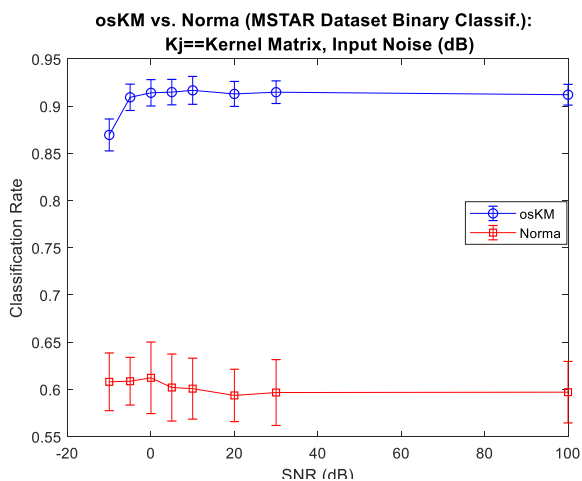

*Figure 2.* Non-Stationary Case: osKM vs. Norma [3] for the varying levels of SNR (dB)

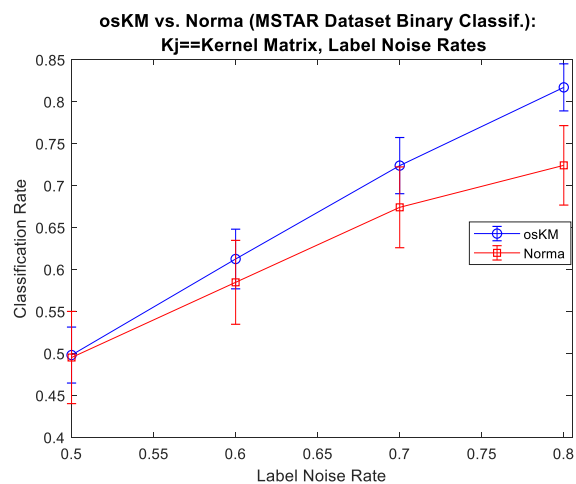

*Figure 3.* Non-Stationary Case: osKM vs. Norma [3] for the varying levels of Label Noise

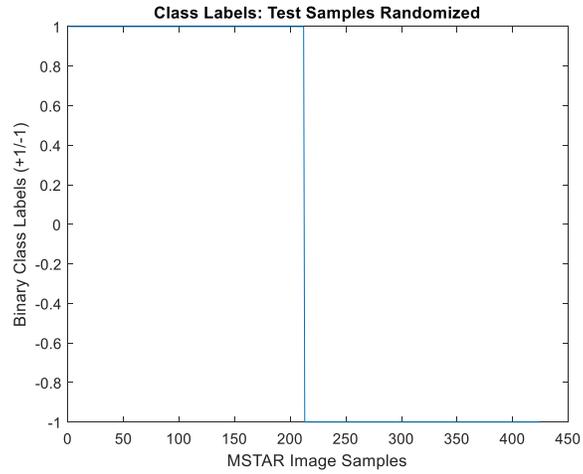

*Figure 1.* Stationary ground-truth labels

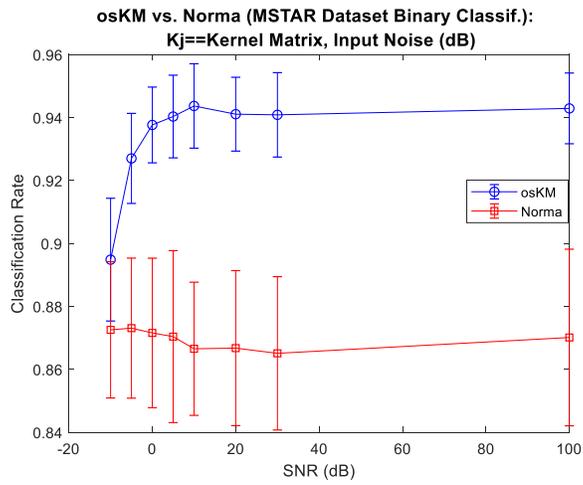

*Figure 2.* Stationary Case: osKM vs. Norma [3] for the varying levels of SNR (dB)

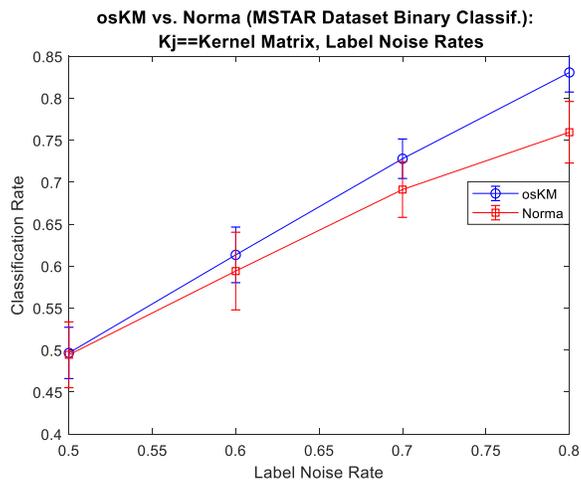

*Figure 3.* Stationary Case: osKM vs. Norma [3] for the varying levels of Label Noise